\documentclass[runningheads]{llncs}

\usepackage{eccv}

\usepackage{eccvabbrv}

\usepackage{graphicx}
\usepackage{booktabs}

\usepackage[table]{xcolor}   
\usepackage{multirow}
\newcommand{\gain}[1]{\textcolor{red}{\scriptsize\,(+#1)}}

\usepackage[accsupp]{axessibility}  

\usepackage{hyperref}

\usepackage{orcidlink}

\begin{document}

\title{Efficient Unified Multimodal Understanding (EUMU): Winning Solution for the\\MUMU Track at the 8th LSVOS Challenge} 
\titlerunning{EUMU: Winning Solution for the MUMU Track}


\author{Dayoung Kil\inst{1}\orcidlink{0000-0002-1341-4734} \and
Seong-heum Kim\inst{1}\thanks{Corresponding author}\orcidlink{0000-0003-2551-0157}}

\authorrunning{D.~Kil and S.-H.~Kim}

\institute{
Department of Intelligent Semiconductors,
\\ Soongsil University, Seoul, Republic of Korea\\
\email{dayoung.k.ssu@gmail.com, seongheum@ssu.ac.kr}
}

\maketitle

\begin{abstract}
The Mobile Unified Multimodal Understanding (MUMU) Challenge requires a single efficient model to jointly perform multi-concept image tagging, open-vocabulary object detection, and image captioning. We present Efficient Unified Multimodal Understanding (EUMU), the winning solution for the MUMU Track of the 8th LSVOS Challenge. EUMU builds on a shared pretrained multimodal model, using its prompt-based capabilities for detection and captioning and training lightweight heads on shared visual features to predict quality, scene, and event tags. Rather than treating the three tasks independently, EUMU applies task-aware inference refinement by reusing task outputs as cross-task cues. For detection, caption cues help recover objects missed by the initial detection. For captioning, detection cues help refine the caption to better reflect the detected objects. For tagging, image statistics refine quality predictions, while caption and detection cues refine scene and event predictions. This design unifies all three tasks within a single model while satisfying the challenge's resource constraints. EUMU contains 239.169M parameters, requires 23.947 GFLOPs, uses 4.5 GB of peak inference memory, and achieves a final challenge score of 17.3409. Code and models are available at \url{https://github.com/Dayoung-Kil/EUMU}.
\end{abstract}

\section{Introduction}
\label{sec:intro}

The 8th Large-scale Video Object Segmentation (LSVOS) Challenge \cite{lsvos26eccv}, held in conjunction with ECCV 2026, includes three main tracks for video object understanding. The MOSEv2 track addresses video object segmentation using LVOS \cite{hong23iccv} and MOSE \cite{ding23iccv,ding25arxiv}. The MeViSv2-Text and MeViSv2-Audio tracks address referring video object segmentation based on MeViS \cite{ding25pami}, with text and audio guidance, respectively. In addition to these tracks, the challenge also introduces the Mobile Unified Multimodal Understanding (MUMU) track \cite{mumu26challenge}, which focuses on efficient, unified multimodal understanding.

MUMU requires a single model to perform three visual understanding tasks under a shared resource budget: multi-concept image tagging, which predicts quality, scene, and event concepts; open-vocabulary object detection, which requires generalization to unseen object categories; and image captioning, which generates a natural-language description of an image. Separate task-specific models are not permitted; all three capabilities must be integrated into a single architecture. The model must also contain no more than 0.5B parameters and use no more than 8 GB of peak inference memory. The challenge therefore evaluates both task performance and resource efficiency.

Meeting these requirements is difficult because the three tasks demand different forms of visual understanding while sharing the same model and computational budget. Conventional pipelines that assign a dedicated model to each task achieve strong per-task accuracy, but their combined parameters, computation, and memory grow with the number of tasks, and each model remains blind to the evidence produced by the others. Recent large multimodal models unify diverse vision-language tasks within a single architecture, yet their scale of several billion parameters far exceeds the 0.5B-parameter and 8 GB memory budget of MUMU. Compact prompt-based models such as Florence-2 fit the budget and already cover detection and captioning, but they offer no native support for multi-concept tagging of quality, scene, and event labels, and their prompt-based inference runs each task in isolation, so the errors of one task are never corrected by the evidence of another. Fine-tuning such a model to add the missing capability would also increase training cost and risk eroding the pretrained abilities on which detection and captioning depend.

To address this challenge, we propose Efficient Unified Multimodal Understanding (EUMU), which combines a frozen shared backbone with inference-time cooperation between tasks. EUMU shares a single pretrained multimodal model for all three tasks. Detection and captioning rely on its prompt-based capabilities, while lightweight heads predict tags from shared visual features. To break the isolation between tasks, EUMU further reuses task outputs as cues to refine predictions across tasks. Extending task coverage without fine-tuning the backbone or enlarging the model, this design allows the predictions of one task to improve another. By combining a shared backbone with task-aware refinement at inference time, EUMU achieves strong performance under the resource constraints and ranks first in the MUMU Track at the 8th LSVOS Challenge 2026.

\section{Method}

\begin{figure}[t]
\begin{center}
    \includegraphics[width=\textwidth]{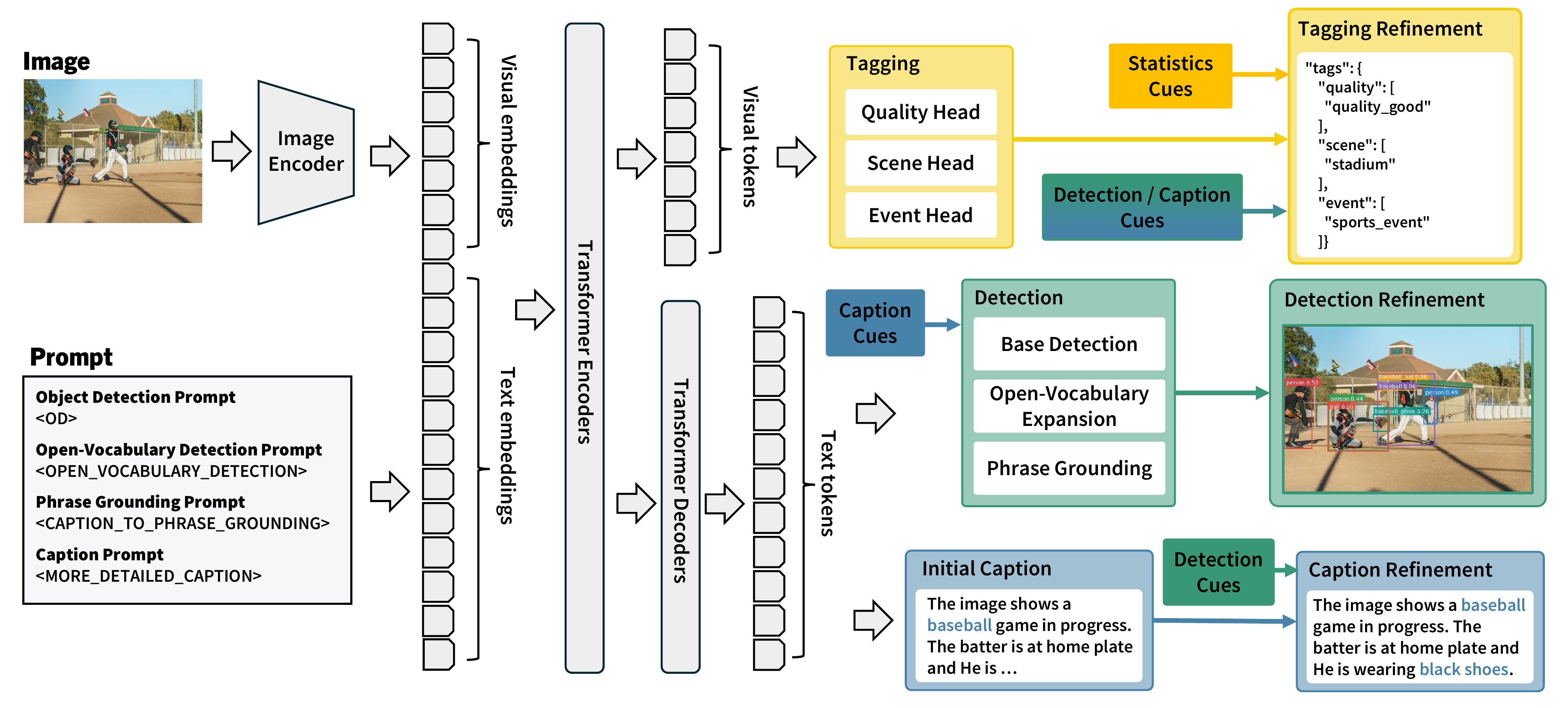}
\end{center}
\caption{\textbf{Overview of EUMU.}
EUMU shares a single pretrained multimodal model~\cite{xiao24cvpr} across multi-concept image tagging, open-vocabulary object detection, and image captioning.
Lightweight tagging heads predict tags from shared visual features, while detection and captioning use prompt-based inference.
Task outputs are selectively reused as cross-task cues for task-aware inference refinement.}
\label{fig:overview}
\end{figure}

Figure~\ref{fig:overview} illustrates the overall architecture of the proposed system for Efficient Unified Multimodal Understanding.
EUMU uses a single pretrained multimodal model to support tagging, detection, and captioning within a unified framework. The input image is encoded into a shared
visual representation through the image encoder and transformer encoder.
Multi-concept tagging is performed directly on these visual features using
lightweight tagging heads for quality, scene, and event prediction. Detection
and captioning, in contrast, use task-specific prompts and the pretrained decoder
without fine-tuning the shared model.

EUMU applies task-aware inference refinement in a directed sequence, reusing task outputs as cross-task cues. EUMU first obtains initial tag predictions, base detections, and a context caption. The context caption provides additional object and phrase cues for open-vocabulary detection and phrase grounding, producing the refined detections. The final detections then guide caption refinement. Finally, image statistics refine the quality predictions, while cues from the final caption and detections refine the scene and event predictions. This process produces the final tags, detections, and caption within a single unified model.

\subsection{Multi-Concept Image Tagging}

For multi-concept image tagging, EUMU introduces three lightweight MLP tagging heads on top of the shared visual features for quality, scene, and event prediction. Each head produces label-wise scores used to obtain the initial tags. Since tagging operates directly on the shared visual representation without using the transformer decoder, EUMU can predict all three types of tags without fine-tuning the pretrained multimodal model.

The initial tag predictions are further refined using additional cues relevant to each label group. Quality labels are closely associated with low-level image characteristics and degradations. We therefore complement the prediction scores with image statistics such as brightness distribution, contrast, gradient, noise, and compression artifacts. For example, brightness statistics are used to refine low-light, underexposure, and overexposure predictions, while contrast-related statistics adjust the corresponding quality labels.
Once the final caption and detections are obtained, scene and event predictions are further refined using their semantic cues. Scene predictions are refined using location cues from the caption together with detected objects, allowing generic predictions such as outdoor to be replaced with more specific scene categories. Event predictions are refined by combining action cues from the caption with object cues from the detections, e.g., stage and microphone for performance or race number and finish line for sports events. Event predictions that are inconsistent with the caption, scene tags, and detection cues are removed.

\subsection{Open-Vocabulary Object Detection}

EUMU augments base object detection with caption-guided open-vocabulary detection and phrase grounding to improve object coverage. It first performs base object detection to obtain bounding boxes and labels. To recover objects missed by the base detection, EUMU extracts additional object and phrase cues from the context caption. These cues are used to detect additional objects through open-vocabulary detection and localize caption phrases through phrase grounding. The context caption therefore serves to expand the detection candidates before final refinement.
Detection candidates from the three stages are merged and refined. The predicted labels are normalized to the evaluation vocabulary, while invalid bounding boxes are removed and non-maximum suppression (NMS) is applied to suppress duplicate predictions. The remaining bounding boxes, labels, and confidence scores form the final detections.

\subsection{Image Captioning}
EUMU uses the \texttt{<MORE\_DETAILED\_CAPTION>} prompt to generate a context caption for detection candidate expansion. After obtaining the final detections, EUMU refines the context caption using detection cues. We compare the detected object labels with the words in the caption and use their overlap as an additional cue for caption refinement.
The refined caption is then post-processed to satisfy the challenge length limits while maintaining sentence completeness. Overly long outputs are truncated to the allowed length, and incomplete sentence endings caused by truncation are removed. We also remove unnecessary whitespace and line breaks and add sentence-ending punctuation when needed. The final caption is subsequently used as a semantic cue for scene and event tagging refinement.

\section{Experiments}

\subsection{Experimental Setup}
\noindent\textbf{Datasets.}
EUMU trains the lightweight tagging heads using the official MUMU training data. To address the sparsity of quality and event labels, we construct additional training samples from the same training split. For quality, we generate synthetic examples by applying degradations such as blur, noise, low light, and overexposure to the training images. For event prediction, we extract event-related cues from the training captions and use them to generate additional event labels. Detection and captioning require no additional training and instead rely on the pretrained prompt-based capabilities of Florence-2-base~\cite{xiao24cvpr}.

For model selection and inference strategy validation, we use task-specific local validation sets. We use KonIQ-10k~\cite{hosu20tip} for quality, Places365~\cite{zhou18pami} for scene, USED-test~\cite{ahmad16mmsys} for event, LVIS v1~\cite{gupta19cvpr} for detection, and the COCO Karpathy validation split~\cite{lin14eccv,karpathy15cvpr} and Flickr30K validation set~\cite{young14tacl} for captioning.

\begin{figure}[t]
\begin{center}
    \includegraphics[width=\textwidth]{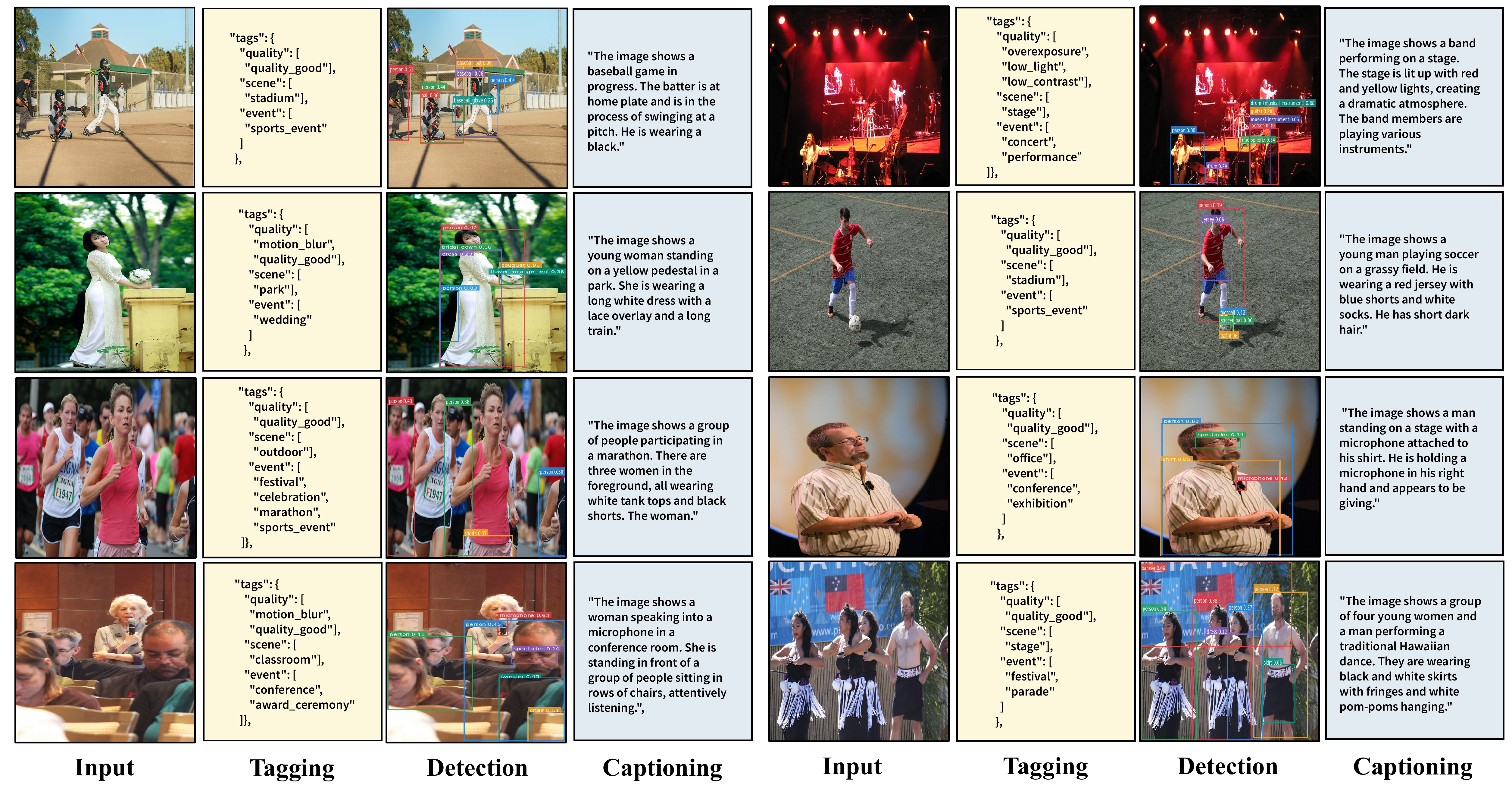}
\end{center}
\caption{\textbf{Qualitative results of EUMU.}
Each example shows the input image together with the final outputs for multi-concept image tagging, open-vocabulary object detection, and image captioning after task-aware inference refinement.
The examples cover diverse image conditions, scenes, and events, illustrating how EUMU captures both low-level quality attributes and high-level semantic information within a single unified model.}
\label{fig:qualitative}
\end{figure}

\vspace{0.2in}
\noindent\textbf{Implementation Details.}
EUMU uses a frozen Florence-2-base~\cite{xiao24cvpr} shared across all three tasks. Only the lightweight tagging heads are trained, while detection and captioning rely on the pretrained prompt-based capabilities without additional fine-tuning.
For tagging, the quality and scene/event heads are trained with AdamW for 6 and 10 epochs, using learning rates of $2\times10^{-4}$ and $1\times10^{-4}$ and batch sizes of 256 and 384, respectively. Both use cosine scheduling, a 0.08 warmup ratio, 0.04 weight decay, and 0.25 dropout.
The final EUMU model contains 239.169M parameters, requires 23.947 GFLOPs, and uses 4.5 GB of peak inference memory.

\noindent\textbf{Evaluation Metrics.}
MUMU evaluates both task performance and model efficiency. Multi-concept image tagging is evaluated using Macro-F1, while open-vocabulary object detection is measured by novel-aware mAP@[0.5:0.95]. Image captioning is evaluated using $0.4$ CIDEr~\cite{vedantam15cvpr} $+\,0.3$ SPICE~\cite{anderson16eccv} $+\,0.3$ CLIPScore~\cite{hessel21emnlp}. The final challenge score combines the three task scores with an efficiency factor.

\subsection{Qualitative Results}
Figure~\ref{fig:qualitative} presents qualitative examples of EUMU across diverse images. For each input, EUMU produces quality, scene, and event tags, object detections, and an image caption within a single unified model. The results show that EUMU captures both image quality characteristics and high-level scene, event, and object information across the three tasks.

\begin{table}[t]
\centering
\caption{\textbf{Effect of task-specific refinement strategies on local validation sets.}
Metrics are reported in their task-specific scales. Values in parentheses indicate the performance gain over the preceding setting, and the best result for each task is shown in bold.}
\label{tab:refinement}
\setlength{\tabcolsep}{10pt}
\renewcommand{\arraystretch}{1.15}
\begin{tabular}{@{}l l c@{}}
\toprule
\textbf{Task} & \textbf{Setting} & \textbf{Local Metric} \\
\midrule
\multirow{2}{*}{\shortstack[l]{\textbf{Tagging}\\{\scriptsize(Macro-F1)}}}
  & Lightweight tagging heads        & 0.4300 \\
  & \enspace+ task-aware refinement  & \textbf{0.4634}\gain{0.0334} \\
\midrule
\multirow{4}{*}{\shortstack[l]{\textbf{Detection}\\{\scriptsize(mAP@[.5:.95])}}}
  & Base object detection            & 0.195 \\
  & \enspace+ phrase grounding       & 0.269\gain{0.074} \\
  & \enspace+ open-vocab.\ expansion & 0.277\gain{0.008} \\
  & \textbf{Full pipeline}           & \textbf{0.334}\gain{0.057} \\
\midrule
\multirow{2}{*}{\shortstack[l]{\textbf{Captioning}\\{\scriptsize(composite)}}}
  & Context caption              & 30.49 \\
  & \enspace+ caption refinement     & \textbf{30.63}\gain{0.14} \\
\bottomrule
\end{tabular}
\end{table}
\subsection{Quantitative Results}
We evaluate EUMU on task-specific local validation sets to analyze the effectiveness of the proposed refinement strategies, and report the final performance on the official MUMU Challenge leaderboard.

\vspace{0.2in}
\noindent\textbf{Effect of Task-Aware Inference Refinement.}
Table~\ref{tab:refinement} reports the effect of the major refinement strategies on the task-specific local validation sets. For tagging, task-aware inference refinement improves the Macro-F1 from 0.4300 to 0.4634. For detection, adding phrase grounding and open-vocabulary expansion increases mAP from 0.195 to 0.334. Caption refinement also improves the composite score from 30.49 to 30.63. These results show that the proposed refinement strategies consistently improve local validation performance without fine-tuning the pretrained multimodal model.

\vspace{0.2in}
\noindent\textbf{Challenge Results.}
Table~\ref{tab:challenge_results} reports the official Codabench leaderboard results for the MUMU Challenge. EUMU achieves the highest final challenge score of 17.3409 and ranks 1st while satisfying the challenge's efficiency constraints.
These results show that EUMU supports unified multimodal understanding efficiently by sharing a pretrained multimodal model across tasks and improving predictions through lightweight tagging heads and task-aware inference refinement.

%

\begin{table}[t]
\centering
\caption{\textbf{Official MUMU Challenge 2026~\cite{mumu26challenge} results and resource usage of EUMU.}
EUMU ranks 1st while satisfying the challenge resource constraints
(Params $\leq$ 0.5B and Peak Memory $\leq$ 8 GB).
The final score is computed as $(0.3A + 0.4B + 0.3C)\times E$,
where $E$ denotes the efficiency factor.}
\label{tab:challenge_results}
\renewcommand{\arraystretch}{1.25}
\begin{subtable}[t]{0.54\textwidth}
\centering
\setlength{\tabcolsep}{5pt}
\resizebox{\linewidth}{!}{%
\begin{tabular}{@{}ccccc@{}}
\toprule
\multirow{2}{*}{\textbf{Rank}} & \multirow{2}{*}{\textbf{Final Score}} & \textbf{Tagging} & \textbf{Detection} & \textbf{Captioning} \\
 & & {\scriptsize(Task A)} & {\scriptsize(Task B)} & {\scriptsize(Task C)} \\
\midrule
\textbf{1st} & \textbf{17.3409} & 27.1808 & 8.2151 & 47.3354 \\
\bottomrule
\end{tabular}}
\caption{Task performance.}
\label{tab:leaderboard}
\end{subtable}\hfill
\begin{subtable}[t]{0.44\textwidth}
\centering
\setlength{\tabcolsep}{5pt}
\resizebox{\linewidth}{!}{%
\begin{tabular}{@{}cccc@{}}
\toprule
\textbf{Params} & \textbf{FLOPs} & \textbf{Peak Mem.} & \textbf{Eff.\ Factor} \\
{\scriptsize(M)} & {\scriptsize(G)} & {\scriptsize(GB)} & {\scriptsize($E$)} \\
\midrule
239.169 & 23.947 & 4.5 & 0.676 \\
\bottomrule
\end{tabular}}
\caption{Resource efficiency.}
\label{tab:resource}
\end{subtable}
\end{table}
\section{Conclusion}

We presented EUMU, the first-place solution to the MUMU Track of the 8th LSVOS Challenge. EUMU handles multi-concept image tagging, open-vocabulary object detection, and image captioning within a single pretrained multimodal backbone. Lightweight tagging heads extend the shared backbone to multi-concept tagging, while detection and captioning retain its pretrained prompt-based capabilities without additional fine-tuning. EUMU further enables cooperation across the three tasks by reusing their outputs as complementary cues for task-aware inference refinement. This design avoids separate task-specific models while improving task predictions within the constrained resource budget. With 239.169M parameters, 23.947 GFLOPs, and 4.5 GB of peak inference memory, EUMU achieves a final challenge score of 17.3409 and ranks first in the MUMU Challenge.


\section*{Acknowledgements}
This work was supported by G-LAMP Program of the National Research Foundation of Korea (NRF) grant funded by the Ministry of Education (No. RS-2025-25441317) and the Ministry of Science and ICT and the National IT Industry Promotion Agency (NIPA) through the Advanced GPU Utilization Support Program (02-26-01-0499). This work was also supported by the National Research Foundation of Korea (NRF) grant funded by the Korea government (MSIT) (RS-2025-16071992). This research was also supported by Korea Institute for Advancement of Technology (KIAT) grant funded by the Korea Government (MOTIE) (RS-2026-25530975, HRD Program for Industrial Innovation).

%
%
\bibliographystyle{splncs04}
\bibliography{main}
\end{document}